\documentclass{article}

\usepackage{arxiv}

\usepackage[utf8]{inputenc} 
\usepackage[T1]{fontenc}    
\usepackage{hyperref}       
\usepackage{url}            
\usepackage{booktabs}       
\usepackage{amsfonts}       
\usepackage{nicefrac}       
\usepackage{microtype}      

\usepackage{lipsum}
\usepackage{multirow}
\usepackage{subcaption}
\usepackage{hhline}
\usepackage{graphicx}
\usepackage{algorithm}
\usepackage{multirow}
\usepackage{subcaption}
\usepackage{algpseudocode}
\usepackage{amsmath}

\usepackage{xcolor}
\definecolor{RoyalBlue}{RGB}{65, 105, 225}

\title{Benchmarking Robustness of Contrastive Learning Models for Medical Image-Report Retrieval}

\author{
    Demetrio Deanda\hspace{3mm} 
    Yuktha Priya Masupalli\hspace{3mm}
    Jeong Yang\hspace{3mm}
    Young Lee\hspace{3mm}
    Zechun Cao\hspace{3mm}
    Gongbo Liang\hspace{3mm}\\ [0.75ex]
    Department of Computational, Engineering, and Mathematical Sciences \\
    Texas A\&M University-San Antonio\\ [0.75ex]
   \texttt{ 
   \{ddean09, ymasu01\}@jaguar.tamu.edu,~\{jyang, ylee, zcao, gliang\}@tamusa.edu}
}

\begin{document}
\maketitle

\begin{abstract}
    Medical images and reports offer invaluable insights into patient health. The heterogeneity and complexity of these data hinder effective analysis. To bridge this gap, we investigate contrastive learning models for cross-domain retrieval, which associates medical images with their corresponding clinical reports. This study benchmarks the robustness of four state-of-the-art contrastive learning models: CLIP, CXR-RePaiR, MedCLIP, and CXR-CLIP. We introduce an occlusion retrieval task to evaluate model performance under varying levels of image corruption. Our findings reveal that all evaluated models are highly sensitive to out-of-distribution data, as evidenced by the proportional decrease in performance with increasing occlusion levels. While MedCLIP exhibits slightly more robustness, its overall performance remains significantly behind CXR-CLIP and CXR-RePaiR. CLIP, trained on a general-purpose dataset, struggles with medical image-report retrieval, highlighting the importance of domain-specific training data. The evaluation of this work suggests that more effort needs to be spent on improving the robustness of these models. By addressing these limitations, we can develop more reliable cross-domain retrieval models for medical applications. 
    \let\thefootnote\relax\footnotetext{This work is accepted to AAAI 2025 Workshop -- the $9^{th}$ International Workshop on Health Intelligence.}

\end{abstract}


\section{Introduction}
The rapid growth of medical data, including images and reports, presents both opportunities and challenges for healthcare professionals. While these data sources offer valuable insights into patient health, their heterogeneity, and complexity can hinder effective analysis and decision-making~\cite{kruse2016challenges}. To bridge this gap, there is a pressing need for AI models capable of jointly understanding both modalities~\cite{pandey2022comprehensive}.

Cross-domain retrieval, which involves establishing connections between data from distinct sources, has the potential to revolutionize medical research and practice. By combining information from multiple domains, healthcare providers can gain a more comprehensive understanding of patient conditions, leading to more accurate diagnoses and personalized treatment plans~\cite{yang2017chexnet,ying2021multi}. Furthermore, cross-domain retrieval can facilitate the discovery of new medical insights by revealing patterns and trends that might otherwise be obscured. In addition, cross-domain retrieval for medical imaging-report can also facilitate the automated generation of medical imaging reports~\cite{endo2021retrieval,wang2022medclip,you2023cxr}.

Contrastive learning has emerged as a promising technique for cross-domain retrieval in medical imaging and reports~\cite{liang2021contrastive,radford2021learning,wang2022medclip}. While neural networks have demonstrated impressive performance in various tasks, such as cyber security~\cite{zulu2024enhancing,liang2023enhancing}, healthcare~\cite{xing2023self,liu2023simulated}, public transportation~\cite{jonnala2025potential,liang2024unveiling}, and astrophysics~\cite{su2020deep, zhang2021multi, lin2022estimating}, modern neural networks are suffering from issues like miscalibration~\cite{guo2017calibration,liang2020imporved,han2024multi}, bias~\cite{wang2020inconsistent}, reliability~\cite{xing2022neural}, and vulnerability to adversarial attacks~\cite{bortsova2021adversarial,zhang2019defense}. To address these limitations, it is crucial to benchmark the robustness of different models.

This paper investigates the robustness of the contrastive learning-based cross-domain retrieval models, including CLIP~\cite{radford2021learning}, CXR-RePaiR~\cite{endo2021retrieval}, MedCLIP~\cite{wang2022medclip}, and CXR-CLIP~\cite{you2023cxr}, for cross-domain retrieval in medical imaging and reports. By establishing benchmarks, we aim to identify strengths, weaknesses, and potential areas for improvement in future research.

\section{Problem Definition}

In the context of medical imaging, cross-domain retrieval involves associating medical images with their corresponding clinical reports~\cite{you2023cxr,endo2021retrieval}. This task is challenging due to the inherent differences in the nature of these data modalities. However, recent advancements in contrastive learning has enabled significant progress in this area~\cite{radford2021learning, liang2021contrastive}.

\subsection{Contrastive learning}
\begin{figure*}[!tb]
\centering
\includegraphics[width=0.975\textwidth]{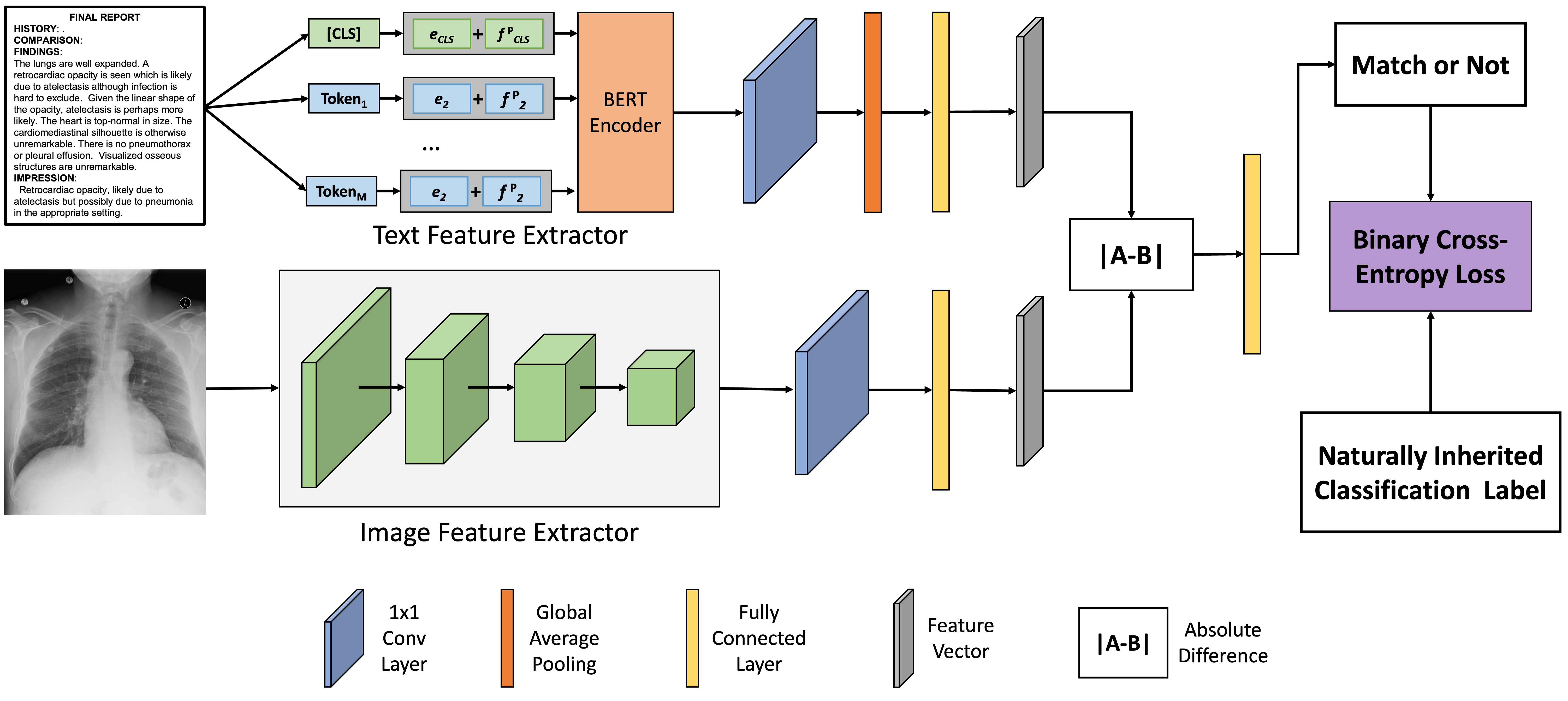}
\caption{Example of a classification-based contrastive learning model. }
\label{fig:architecture}
\end{figure*}

Contrastive learning is a technique that learns general data representations by comparing similar and dissimilar samples. In the imaging domain, Siamese networks, composed of two identical subnetworks, are commonly used for this purpose~\cite{taigman2014deepface}. These networks process pairs of images and output feature vectors, which are then compared using a loss function like triplet loss~\cite{schroff2015facenet} or contrastive loss~\cite{chen2020simple, he2020momentum}. Alternatively, a binary cross-entropy loss can be used to train Siamese networks as a binary classification problem~\cite{koch2015siamese, yang2020mscnn}.

To jointly understand medical images and their corresponding reports, a Siamese-style network with distinct subnetworks can be employed. Given a dataset $X$ of image-text pairs $\{(I_j, T_k)\}$, where $I_j \in \mathcal{I}$ is an image and $T_k \in \mathcal{T}$, $\mathcal{I}$ and $\mathcal{T}$ denoting a set of images and a set of textual reports, respectively. If $j=k$, the image and textural report are associated (i.e., the image and report are matching). The goal of contrastive learning is to learn a model $h(\cdot)$ that pulls the embeddings of matching image-text pairs closer in the feature space and pushes those of non-matching pairs further apart.

The model $h(\cdot)$ typically consists of two branches: $h_i(\cdot)$ for processing images and $h_t(\cdot)$ for processing text. The image branch often employs a convolutional neural network (CNN) or a Vision Transformer (ViT), such as ResNet~\cite{he2016deep} or Vision Transformer~\cite{dosovitskiy2020image}, to extract visual features $v_i$.  The text branch, usually is a large language model (LLM) like BERT~\cite{kenton2019bert} or RoBERTa~\cite{liu2019roberta}, extracts textural features $v_t$. These features are then projected to the same space and compared using contrastive losses, such as triplet loss, or classified using a binary classifier.

\subsection{Contrastive Learning for Cross-Domain Retrieval}
Contrastive learning is a powerful technique for training cross-domain retrieval models, which can be employed in two ways: similarity comparison and binary classification.

Contrastive learning brings together samples from the same group (e.g., a matching image and report) in the feature space, while simultaneously pushing apart samples from different groups (e.g., non-matching pairs). This enables a straightforward approach to retrieval: by comparing the embeddings similarity between images and reports in the feature space. A higher similarity score indicates a stronger match~\cite{endo2021retrieval,radford2021learning}.

Alternatively, when a contrastive learning model is trained as a binary classifier, the classification model itself can be directly utilized for retrieval. As illustrated in Figure~\ref{fig:architecture}, the model processes an image $I_j$ through the image processing branch $h_i(\cdot)$ and a text report $T_k$ though the text processing branch $h_t(\cdot)$. The absolute difference between the resulting embedding vectors, $v_i$ and $v_t$,  is fed into a shallow classification model to determine whether the pair is a match. The classification probability can then be used as a retrieval score~\cite{liang2021contrastive}.

\section{Method}
This work investigates the robustness of four contrastive learning models applied to medical image-report retrieval tasks. Given a query image, the objective is to retrieve the most relevant report. This section outlines the detailed evaluation methodology.

\subsection{Robustness Evaluation}

To assess the robustness of the pre-trained contrastive learning-based cross-domain retrieval methods, we introduced an occlusion retrieval task. During evaluation, we systematically occluded a portion ($p$) of the image ($p = \{0\%, 0.25\%, 1\%, 4\%, 9\%, 25\%, 49\%, 81\%\}$) at random locations, generating out-of-distribution data for the pre-trained models. These occluded images were then used as input to the models for retrieval tasks. 

To evaluate the robustness of the models, we calculated Recall@$k$, a metric that measures the proportion of relevant items retrieved within the top $k$ results. We varied the value of $k$ ($\{5, 10, 20, 30, 50, 100\}$) to assess performance at different retrieval depths. Recall@$k$ is calculated as follows:
\begin{equation}
    Recall@k = \frac{\text{\# of relevant items retrieved in top k}}{\text{Total \# of relevant items}}.
\end{equation}
Ideally, a robust model should exhibit similar $Recall@k$ values across different occlusion levels, especially for smaller occlusion percentages $p$. 

Algorithm~\ref{alg:occ_test} provides a detailed description of occlusion retrieval with a specific occlusion ratio.

\begin{algorithm*}[!tb]
    \caption{Occlusion Retrieval Test for a Pre-Trained Image-Text Retrieval Model}
    \footnotesize
    \begin{algorithmic}
    \Require Pre-Trained Image-Text Retrieval Model $h(\cdot)$, Chest X-Ray Imaging Set $\mathcal{I}$ with $M$ images, Textural Report Set $\mathcal{T}$ with $N$ reports, Occlusion ratio $p$ indicating the percentage pixel will be blocked, Constant $k$ for calculating Recall@$K$
        \State
        \State $total\_correct \gets 0$ \Comment{The number of relevant report retrieved within the top $k$ results}
        
        \State
        \For{$m \gets 0$ to $M-1$} \Comment{For every image}
            \State $i \gets \mathcal{I}[m]$ \Comment{Get the $m^{th}$ chest x-ray image}
            \State $S \gets [~]$ \Comment{An empty array holding the matching score between an image and all reports}
            
            \State
            \For{$n \gets o$ to $N-1$} \Comment{For every report}
                \State $t \gets \mathcal{T}[n]$ \Comment{Get the $n^{th}$ report}
                
                \State $i_o \gets \phi(i, p)$ \Comment{Generate the occlusion version of $i$ by random block a $p$\% of pixels}

                \State $s \gets h(i_o, t)$ \Comment{A score indicates the degree of matching between $i_o$ and $t$,}
                \State \Comment{a higher value indicates a better match}

                \State $S$.append($[t, s]$) \Comment{Append the report and matching score to the array $S$}
            \EndFor

            \State            
            \State $S$.sort() \Comment{Sort $S$ according $s$ in an descending order}
            
            \State
            \If{the corresponding report in the top $k$ items in $S$}  \Comment{If the matched report is in the top}
            
            \Comment{$k$ retrieved items}
                \State $total\_correct \gets total\_correct + 1$ \Comment{Increase the number by 1}

            \EndIf
            
        \State
        \EndFor

        \State
        \State $recall \gets total\_correct / M $ \Comment{Calculate Recall@$K$}

        \State
        \State \Return $recall$ \Comment{Return Recall@$K$}
    
    \end{algorithmic}
    \label{alg:occ_test}
\end{algorithm*}

\subsection{Cross-Domain Retrieval Models}
This work evaluates four contrastive learning-based models for cross-domain retrieval tasks: CLIP~\cite{radford2021learning}, CXR-RePaiR~\cite{endo2021retrieval}, MedCLIP~\cite{wang2022medclip}, and CXR-CLIP~\cite{you2023cxr}.

\subsubsection{CLIP (Contrastive Language-Image Pre-training)} This neural network learns a shared feature space for images and text. Trained on a massive dataset of image-text pairs, CLIP maximizes similarity between semantically related pairs while minimizing it for unrelated ones. This allows CLIP to understand the connection between visual and textual information, enabling tasks like image classification and zero-shot learning. In our work, we leverage CLIP's learned embeddings for image-text retrieval by calculating cosine similarity between image and text embeddings generated by the pre-trained model.

\subsubsection{CXR-RePaiR (Contrastive X-ray-Report Pair Retrieval)} This method generates chest X-ray reports through a retrieval-based fashion. It fine-tuned CLIP on the MIMIC-CXR~\cite{johnson2019mimic} dataset for report-level or sentence-level retrieval. Report-level retrieval selects the entire best-matching report from the candidate set, while sentence-level retrieval constructs a new report by selecting sentences from multiple reports. For consistency with other methods, we employ the report-level retrieval in this work, calculating cosine similarity between query image embeddings and textual report embeddings generated by a CLIP model that was initialized with CXR-RePaiR weights (available on their official GitHub repository\footnote{https://github.com/rajpurkarlab/CXR-RePaiR}).

\subsubsection{MedCLIP} This neural network model is jointly trained on medical images and their corresponding text reports. Unlike previous methods, MedCLIP utilizes unpaired data, reducing the need for large amount of paired data. Designed as a general-purpose medical imaging model, MedCLIP may perform various tasks like zero-shot learning, supervised classification, and image-text retrieval. Here, we focus on its image-text retrieval capabilities. We leverage publicly available code and pre-trained weights from the official MedCLIP GitHub repository\footnote{https://github.com/RyanWangZf/MedCLIP} in this study.

\subsubsection{CXR-CLIP} Similar to MedCLIP, CXR-CLIP aims to train a general-purpose image-text model using limited data. However, instead of unpaired data, CXR-CLIP leverages Large Language Models (LLMs) to expand image-label pairs into natural language descriptions. Additionally, it utilizes multiple images and report sections for contrastive learning. To effectively learn image and textual features, CXR-CLIP introduces two novel loss functions: ICL and TCL. ICL focuses on learning study-level characteristics of medical images, while TCL focuses on learning report-level characteristics. Pre-trained CXR-CLIP models can perform both zero-shot learning and image-text retrieval. We evaluate CXR-CLIP's image-text retrieval capabilities using the official code and pre-trained model available on CXR-CLIP's official GitHub repository\footnote{https://github.com/Soombit-ai/cxr-clip}.

\subsection{MIMIC-CXR Dataset}
\begin{figure*}
  \begin{subfigure}[b]{0.27\textwidth}
    ~~~\includegraphics[width=.73\textwidth]{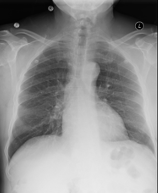}~~~~~
    \vspace{5.25mm}
  \end{subfigure}
  \begin{subfigure}[b]{0.73\textwidth}
    \footnotesize
    \textbf{HISTORY:} \_\_\_-year-old female with chest pain. \\
    
    \textbf{COMPARISON:} Comparison is made with chest radiographs from \_\_\_. \\
    
    \textbf{FINDINGS:} The lungs are well expanded. A retrocardiac opacity is seen which is likely due to atelectasis although infection is hard to exclude. Given the linear shape of the opacity, atelectasis is perhaps more likely. The heart is top-normal in size. The cardiomediastinal silhouette is otherwise unremarkable. There is no pneumothorax or pleural effusion. Visualized osseous structures are unremarkable. \\
    
    \textbf{IMPRESSION:} Retrocardiac opacity, likely due to atelectasis but possibly due to pneumonia in the appropriate setting. \\
  \end{subfigure}
  \caption{Example of a chest x-ray (left) with the radiology report (right) from the MIMIC-CXR dataset~\cite{johnson2019mimic}.}
  \label{fig:data_example}  
\end{figure*}

The MIMIC-CXR dataset~\cite{johnson2019mimic} is a dataset that widely used for contrastive learning and image-text retrieval in the medical domain. The dataset contains 227,835 radiographic studies from 64,588 patients, encompassing 368,948 chest X-rays and their corresponding radiology reports. The dataset also provides 14 labels (13 for abnormalities and one for normal cases) derived from radiology reports using NLP tools like NegBio~\cite{peng2018negbio} and CheXpert~\cite{irvin2019chexpert}.

The official validation set includes 2,991 imaging studies, each containing one or more chest X-rays paired with a single textual report (e.g. Figure~\ref{fig:data_example} ). Each report is divided into sections such as History, Comparison, Findings, and Impression. To ensure data quality, we filtered out reports missing the Findings or Impression sections, resulting in a final validation set of 994 studies with 1,770 X-rays. This filtered dataset is, then, used in our experiments.

\section{Result}
\subsection{Cross-Domain Retrieval}
Table~\ref{tab:result} presents the occlusion retrieval results of the four evaluated models for various occlusion percentages. \textbf{Bold text} highlight the best performance for each occlusion ratio and recall threshold. \textcolor{blue}{Blue text} indicates the second-best performance, while \textcolor{red}{red text} denotes the worst performance. 

\begin{table*}[!tb]
    \setlength{\tabcolsep}{2.25pt}
    \centering
    \begin{tabular}{c|c||c|c|c|c|c|c|c|c||c}
    \hline\hline
         \multirow{2}{*}{} & \multirow{2}{*}{\textbf{Method}} & \multicolumn{8}{c||}{\textbf{Occlusion Area in Percentage}} & \textbf{Random} \\\cline{3-10}
         & & $\bf{0.00\%}$ & $\bf{0.25\%}$ & $\bf{1.00\%}$ & $\bf{4.00\%}$ & $\bf{9.00\%}$ & $\bf{25.00\%}$ & $\bf{49.00\%}$ & $\bf{81.00\%}$ & \textbf{Performance} \\\hline\hline
        
        \multirow{4}{*}{\textbf{Recall @ 5}} & \textbf{CLIP} & \textcolor{red}{$0.57$} & \textcolor{red}{$0.45$} & \textcolor{red}{$0.51$} & \textcolor{red}{$0.40$} & \textcolor{red}{$0.51$} & \textcolor{red}{$0.40$} & \textcolor{red}{$0.62$} & $\bf{0.45}$ & \multirow{4}{*}{0.50}\\\cline{2-10}
        & \textbf{CXR-RePaiR} & \textcolor{blue}{$14.76$} & \textcolor{blue}{$14.54$} & \textcolor{blue}{$13.69$} & \textcolor{blue}{$13.35$} & \textcolor{blue}{$10.24$} & \textcolor{blue}{$5.43$} & \textcolor{blue}{$1.36$} & $\bf{0.45}$ \\\cline{2-10}
        & \textbf{CXR-CLIP} & $\bf{48.56}$ & $\bf{47.26}$ & $\bf{46.92}$ & $\bf{43.47}$ & $\bf{37.42}$ & $\bf{20.41}$ & $\bf{5.94}$ & \textcolor{red}{$0.34$}\\\cline{2-10}
        & \textbf{MedCLIP} & $1.19$ & $1.24$ & $1.30$ & $1.19$ & $1.02$ & $0.73$ & $0.62$ & \textcolor{blue}{$0.40$} \\\hline\hline

        \multirow{4}{*}{\textbf{Recall @ 10}} & \textbf{CLIP} & \textcolor{red}{$1.07$} & \textcolor{red}{$1.07$} & \textcolor{red}{$0.90$} & \textcolor{red}{$0.90$}  & \textcolor{red}{$0.85$} & \textcolor{red}{$0.79$} & \textcolor{red}{$1.07$} & $0.85$ & \multirow{4}{*}{0.99}\\\cline{2-10}
        & \textbf{CXR-RePaiR} & \textcolor{blue}{$23.47$} & \textcolor{blue}{$22.45$} & \textcolor{blue}{$22.45$} & \textcolor{blue}{$20.31$} & \textcolor{blue}{$16.86$} & \textcolor{blue}{$8.54$} & \textcolor{blue}{$2.83$} & \textcolor{red}{$0.74$} \\\cline{2-10}
        & \textbf{CXR-CLIP} & $\bf{58.68}$ & $\bf{57.66}$ & $\bf{58.06}$ & $\bf{52.63}$ & $\bf{46.24}$ & $\bf{27.93}$ & $\bf{9.84}$ & $\bf{1.13}$\\\cline{2-10}
        & \textbf{MedCLIP} & $2.37$ & $2.43$ & $2.54$ & $2.77$ & $2.09$ & $1.69$ & $1.36$ & \textcolor{blue}{$0.96$} \\\hline\hline

        \multirow{4}{*}{\textbf{Recall @ 20}} & \textbf{CLIP} & \textcolor{red}{$2.04$} & \textcolor{red}{$1.98$} & \textcolor{red}{$1.81$} & \textcolor{red}{$1.75$} & \textcolor{red}{$1.75$} & \textcolor{red}{$1.75$} & \textcolor{red}{$1.81$} & \textcolor{blue}{$1.70$} & \multirow{4}{*}{2.01} \\\cline{2-10}
        & \textbf{CXR-RePaiR} & \textcolor{blue}{$34.05$} & \textcolor{blue}{$34.05$} & \textcolor{blue}{$33.37$} & \textcolor{blue}{$29.36$} &\textcolor{blue}{$25.34$} & \textcolor{blue}{$13.63$} & \textcolor{blue}{$4.47$} & \textcolor{red}{$1.36$} \\\cline{2-10}
        & \textbf{CXR-CLIP} & $\bf{67.55}$ & $\bf{67.44}$ & $\bf{65.86}$ & $\bf{61.56}$ & $\bf{56.08}$ & $\bf{38.33}$ & $\bf{15.43}$ & $\bf{1.98}$\\\cline{2-10}
        & \textbf{MedCLIP} & $4.46$ & $4.63$ & $4.24$ & $3.95$ & $4.12$ & $3.33$ & $2.71$ & $1.64$ \\\hline\hline

        \multirow{4}{*}{\textbf{Recall @ 30}} & \textbf{CLIP} & \textcolor{red}{$3.00$} & \textcolor{red}{$3.00$} & \textcolor{red}{$2.71$} & \textcolor{red}{$2.71$} & \textcolor{red}{$2.71$} & \textcolor{red}{$2.83$} & \textcolor{red}{$3.11$} & $2.38$ & \multirow{4}{*}{3.02} \\\cline{2-10}
        & \textbf{CXR-RePaiR} & \textcolor{blue}{$40.61$} & \textcolor{blue}{$39.93$} & \textcolor{blue}{$38.97$} & \textcolor{blue}{$36.26$} & \textcolor{blue}{$32.24$} & \textcolor{blue}{$17.65$} & \textcolor{blue}{$5.60$} & \textcolor{red}{$1.87$} \\\cline{2-10}            
       & \textbf{CXR-CLIP} & $\bf{73.21}$ & $\bf{71.91}$ & $\bf{71.34}$ & $\bf{67.55}$ & $\bf{61.39}$ & $\bf{43.75}$ & $\bf{19.45}$ & $\bf{3.00}$\\\cline{2-10}
        & \textbf{MedCLIP} & $5.82$ & $6.10$ & $6.05$ & $5.59$ & $5.88$ & $4.75$ & $3.73$ & \textcolor{blue}{$2.66$} \\\hline\hline

        \multirow{4}{*}{\textbf{Recall @ 50}} & \textbf{CLIP} & \textcolor{red}{$5.20$} & \textcolor{red}{$4.81$} & \textcolor{red}{$4.36$} & \textcolor{red}{$4.58$} & \textcolor{red}{$4.24$} & \textcolor{red}{$4.36$} & \textcolor{red}{$5.54$} & $4.13$ & \multirow{4}{*}{5.03} \\\cline{2-10}
        & \textbf{CXR-RePaiR} & \textcolor{blue}{$49.66$} & \textcolor{blue}{$49.21$} & \textcolor{blue}{$48.59$} & \textcolor{blue}{$45.31$} & \textcolor{blue}{$41.57$} & \textcolor{blue}{$24.77$} & \textcolor{blue}{$9.39$} & \textcolor{red}{$3.96$} \\\cline{2-10}
        & \textbf{CXR-CLIP} & $\bf{79.48}$ & $\bf{78.58}$ & $\bf{78.58}$ & $\bf{74.51}$ & $\bf{67.83}$ & $\bf{52.80}$ & $\bf{28.21}$ & $\bf{5.48}$\\\cline{2-10}
        & \textbf{MedCLIP} & $9.21$ & $8.98$ & $9.38$ & $8.47$ & $8.93$ & $7.51$ & $6.21$ & \textcolor{blue}{$4.52$} \\\hline\hline

        \multirow{4}{*}{\textbf{Recall @ 100}} & \textbf{CLIP} & \textcolor{red}{$10.07$} & \textcolor{red}{$9.39$} & \textcolor{red}{$9.11$} & \textcolor{red}{$9.56$} & \textcolor{red}{$9.11$} & \textcolor{red}{$9.56$} & \textcolor{red}{$9.45$} & $8.60$ & \multirow{4}{*}{9.94} \\\cline{2-10}
        & \textbf{CXR-RePaiR} & \textcolor{blue}{$64.03$} & \textcolor{blue}{$64.14$} & \textcolor{blue}{$62.73$} & \textcolor{blue}{$59.39$} & \textcolor{blue}{$56.17$} & \textcolor{blue}{$37.84$} & \textcolor{blue}{$18.27$} & \textcolor{red}{$8.54$} \\\cline{2-10}
        & \textbf{CXR-CLIP} & $\bf{88.19}$ & $\bf{87.73}$ & $\bf{87.51}$ & $\bf{84.17}$ & $\bf{79.54}$ & $\bf{66.76}$ & $\bf{40.53}$ & $\bf{11.42}$\\\cline{2-10}
        & \textbf{MedCLIP} & $16.50$ & $16.61$ & $17.23$ & $16.16$ & $16.27$ & $14.29$ & $12.82$ & \textcolor{blue}{$9.83$} \\\hline\hline
        
    \end{tabular}
    \caption{Occlusion retrieval results of all the models at various occlusion ratio (from 0\% to 81\%)}
    \label{tab:result}
\end{table*}

The table reveals that CXR-CLIP consistently achieves the best performance across most occlusion ratios and recall thresholds, except for the 81\% occlusion level for Recall@5. CXR-RePaiR consistently achieves the second-best performance for all occlusion ratios, except for the 81\% occlusion level. MedCLIP generally ranks third, but it achieves the second-best performance five times at the 81\% occlusion level across six different recall thresholds. CLIP consistently performs the worst, with most results aligning with random performance. 

While CLIP's poor performance is expected due to its training on natural images, MedCLIP's relatively weaker performance is surprising, given its training on medical data. However, this aligns with the performance trends reported in the MedCLIP paper, where MedCLIP outperforms CLIP by approximately two times~\cite{wang2022medclip}. We believe MedCLIP's weaker retrieval performance stems from its integration of unpaired images, texts, and labels using a rule-based labeler, which may hinder the model's ability to accurately associate images with their corresponding reports due to the decoupling of image-text pairs.

\subsection{Robustness Analysis}

\begin{figure}
    \centering
    \begin{subfigure}[b]{0.995\textwidth}
        \includegraphics[width=.995\textwidth]{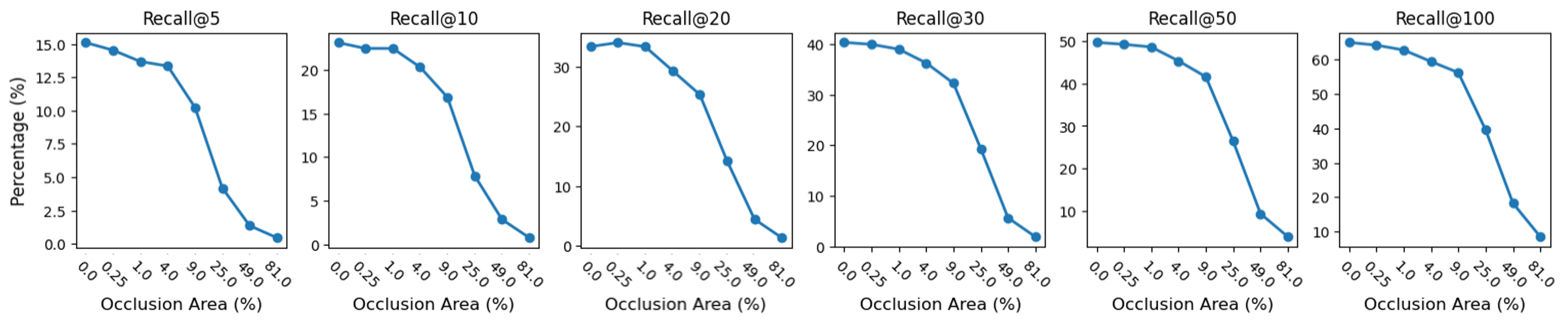}
    \end{subfigure}

    \begin{subfigure}[b]{0.995\textwidth}
        \includegraphics[width=.995\textwidth]{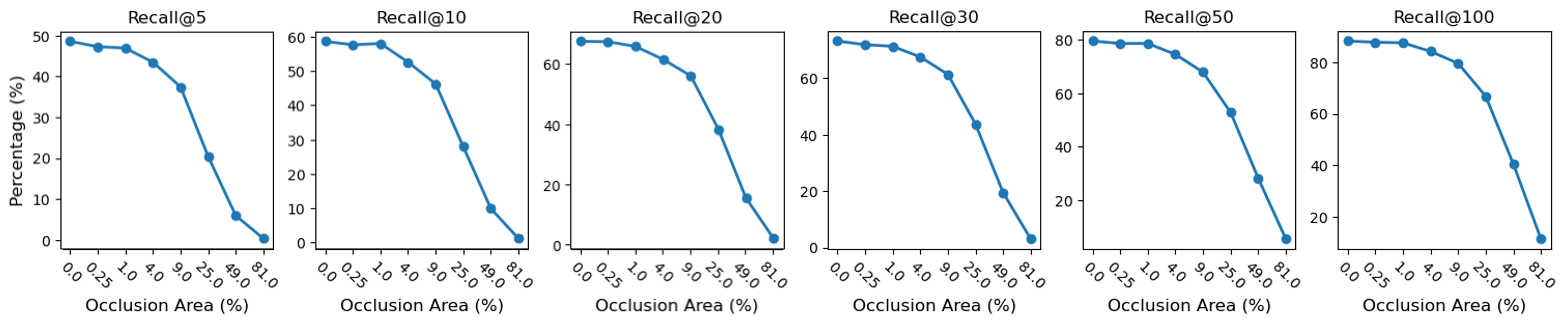}
    \end{subfigure}

    \begin{subfigure}[b]{0.995\textwidth}
        \includegraphics[width=.995\textwidth]{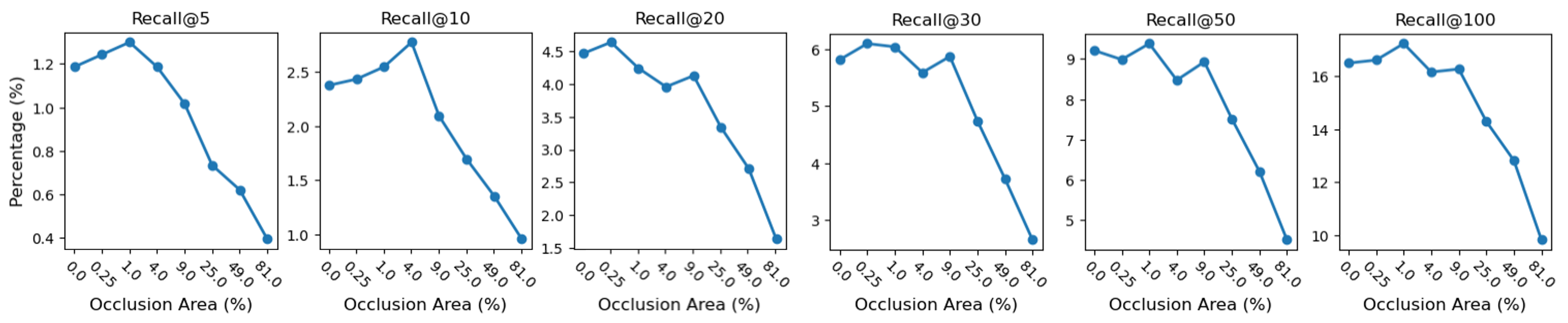}
    \end{subfigure}

    \caption{Robustness testing result for CXR-RePaiR (top), CXR-CLIP (middle), and ViT-based MedCLIP (bottom).}
    \label{fig:robust}
\end{figure}

Figure~\ref{fig:robust} visualizes the performance of CXR-RePaiR (Figure~\ref{fig:robust} top), CXR-CLIP (Figure~\ref{fig:robust} middle), and MedCLIP (Figure~\ref{fig:robust} bottom), respectively. All three models exhibit a decrease in performance as the image occlusion percentage increases. The performance degradation is generally proportional to the occlusion level, with MedCLIP showing a slightly slower decline (approximately 20\%) compared to the other two models. This near-proportional performance decrease suggests that none of the models are robust to handle occluded or out-of-distribution data.

Between CXR-RePaiR and CXR-CLIP, CXR-RePaiR shows a slightly steeper decline in performance, indicating lower robustness compared to CXR-CLIP.

While MedCLIP exhibits a weaker overall retrieval performance, its slower decline in performance suggests potential robustness. Especially for low occlusion levels (less than 4\%), slight occlusions may even improve MedCLIP's retrieval performance. We hypothesize that this is due to the model's training on unpaired images, texts, and labels. Slight occlusions may act as a form of noise reduction, smoothing out potential overfitting and improving generalization.

\section{Conclusion}
This study investigates the robustness of contrastive learning-based cross-domain retrieval models for medical image-report retrieval tasks. By introducing an occlusion retrieval task, we assessed the performance of CLIP, CXR-RePaiR, MedCLIP, and CXR-CLIP under varying levels of image corruption.

Our findings indicate that CXR-CLIP consistently outperforms the other models, demonstrating superior retrieval performance. CXR-RePaiR exhibits the second-best performance, while MedCLIP, despite its potential, shows a weaker overall performance, especially in the presence of significant occlusions. CLIP, trained on a general-purpose dataset, struggles with medical image-report retrieval, highlighting the importance of domain-specific training data.

However, all the evaluated models are extremely sensitive to out-of-distribution data, as shown by the proportional decrease in performance with increasing occlusion percentages. While MedCLIP might exhibit slightly more robustness, its overall performance remains behind CXR-CLIP and CXR-RePaiR. We hypothesize that this is due to its training on unpaired images, texts, and labels. Slight occlusions may act as a form of noise reduction, improving generalization. However, the decoupling of image-text pairs in the unpaired training setting may limit the model's ability to accurately associate images with their corresponding reports.

Future research should explore techniques to improve the robustness of contrastive learning models. Additionally, investigating the impact of different types of data augmentation and architectural modifications on model performance is crucial. By addressing these limitations, we can develop more robust and reliable cross-domain retrieval models for medical applications.

\section*{Acknowledgment}
 This material is based upon work supported by the National Science Foundation’s Grant No. 2334243. Any opinions, findings, and conclusions or recommendations expressed in this material are those of the author(s) and do not necessarily reflect the views of National Science Foundation.

\bibliographystyle{IEEEtran}  
\bibliography{bibfile}

\end{document}